\documentclass[11pt]{article}

\usepackage[preprint]{acl}

\usepackage{times}
\usepackage{latexsym}
\usepackage{algorithm}          
\usepackage{algpseudocode}      
\usepackage{amsmath}            
\usepackage{amssymb}            
\usepackage{amsfonts}           
\usepackage{dsfont}             
\usepackage{booktabs}
\usepackage{multirow}
\usepackage{array}
\usepackage[table]{xcolor} 
\usepackage{booktabs}      
\usepackage{multirow}      
\usepackage{tabularx}

\AtBeginDocument{
  \setlength{\abovedisplayskip}{5pt}
  \setlength{\belowdisplayskip}{5pt}
}

\usepackage[T1]{fontenc}
\usepackage[utf8]{inputenc}

\DeclareTextCommand{\vs}{T1}{\v s}
\DeclareTextCommand{\vc}{T1}{\v c}

\usepackage[T1]{fontenc}

\usepackage[utf8]{inputenc}

\usepackage{microtype}

\usepackage{inconsolata}

\usepackage{graphicx}
\usepackage{caption}
\usepackage{subcaption}

\usepackage{amsmath}

\title{Graph Reasoning Paradigm: Structured and Symbolic Reasoning with Topology-Aware Reinforcement Learning for Large Language Models}

\author{
  \textbf{Runxuan Liu\textsuperscript{1}},
  \textbf{Xianhao Ou\textsuperscript{1}},
  \textbf{Xinyan Ma\textsuperscript{1}},
  \textbf{Jiyuan Wang\textsuperscript{1}},
  \textbf{Jiafeng Liang\textsuperscript{1,\textdagger}},
  \textbf{Jiaqi Li\textsuperscript{2,3,\textdagger}},\\
  \textbf{Tao He\textsuperscript{1}},
  \textbf{Zheng Chu\textsuperscript{1}},
  \textbf{Rongchuan Mu\textsuperscript{1}},
  \textbf{Zekun Wang\textsuperscript{1}},
  \textbf{Baoxin Wang\textsuperscript{2}},\\
  \textbf{Dayong Wu\textsuperscript{2}},
  \textbf{Ming Liu\textsuperscript{1,4,\textdagger}},
  \textbf{Shijin Wang\textsuperscript{2}},
  \textbf{Guoping Hu\textsuperscript{2}},
  \textbf{Bing Qin\textsuperscript{1,4}}\\
  \textsuperscript{1}Harbin Institute of Technology, Harbin, China\\
  \textsuperscript{2}State Key Laboratory of Cognitive Intelligence, iFLYTEK Research, China\\
  \textsuperscript{3}Tianjin Normal University, Tianjin, China
  \textsuperscript{4}Pengcheng Laboratory, Shenzhen, China\\[-0.3em]
  \small\quad\texttt{\{rxliu, xhou, xyma, jywang, jfliang, mliu, qinb\}@ir.hit.edu.cn}\\[-0.3em]
  \small\textdagger\ Corresponding authors.\\
}

\begin{document}
\maketitle
\begin{abstract}

Long Chain-of-Thought (LCoT), achieved by Reinforcement Learning with Verifiable Rewards (RLVR), has proven effective in enhancing the reasoning capabilities of Large Language Models (LLMs).
However, reasoning in current LLMs is primarily generated as plain text, where performing semantic evaluation on such unstructured data creates a computational bottleneck during training.
Despite RLVR-based optimization, existing methods still suffer from coarse-grained supervision, reward hacking, high training costs, and poor generalization.
To address these issues, we propose the Graph Reasoning Paradigm (\textbf{GRP}), which realizes structured and symbolic reasoning, implemented via graph-structured representations with step-level cognitive labels. 
Building upon GRP, we further design Process-Aware Stratified Clipping Group Relative Policy Optimization (\textbf{PASC-GRPO}), which leverages structured evaluation to replace semantic evaluation, achieves process-aware verification through graph-structured outcome rewards, and mitigates reward hacking via stratified clipping advantage estimation.
Experiments demonstrate significant improvements across mathematical reasoning and code generation tasks. Data, models, and code will be released later.

\end{abstract}

\section{Introduction}
\begin{figure}[t]
    \centering
    \includegraphics[width=1 \linewidth]{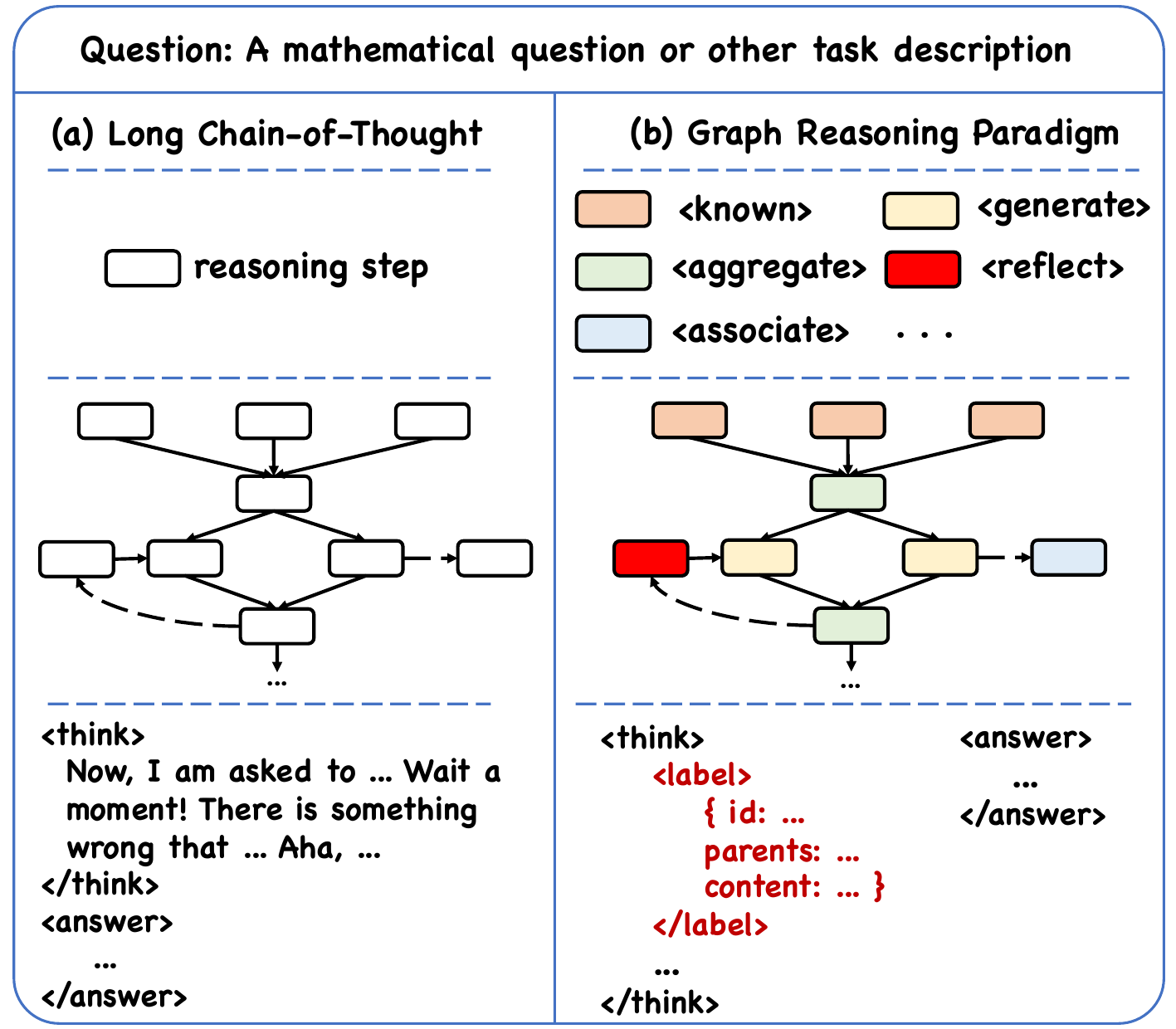}
    \caption{Comparison between traditional Long Chain-of-Thought and Graph Reasoning Paradigm.}
    \label{fig:example}
\end{figure}

Long chain-of-thought (LCoT) has been proven effective for eliciting the reasoning capabilities of Large Language Models (LLMs), particularly when combined with reinforcement learning with verifiable rewards (RLVR)~\cite{ElKishky2024OpenAIOS,Chen2025TowardsRE,He2025BreakingTR}. 
Such approaches have been applied to mathematical reasoning~\cite{Zhou2024SelfDiscoverLL,Moshkov2025AIMO2WS} and code generation~\cite{ElKishky2025CompetitivePW,Yu2025Z1ET}, which are representative complex reasoning tasks and have become common benchmarks for evaluating the reasoning abilities of LLMs.


However, current LLMs predominantly produce plain-text outputs. Although complex reasoning behaviors such as reflection and Aha moments have emerged \cite{Yang2025UnderstandingAM}, the reliability of the reasoning process remains limited \cite{Shojaee2025TheIO}.
Recently, several reinforcement learning (RL) approaches have attempted to address this issue. However, outcome-based reward methods provide only coarse-grained supervision, offering little semantic control over the reasoning process \cite{DeepSeekAI2025DeepSeekR1IR}. Although process-based reward methods offer semantic control over the reasoning process, they often suffer from high training costs and poor generalization \cite{Yuan2024FreePR}.
In summary, semantic evaluation is inherently difficult. Since LLMs output plain text, we are forced to perform semantic assessments, which creates a vicious cycle.

Inspired by the structured cognitive mechanisms of human cognition \cite{George2021ClonestructuredGR}, we propose to replace plain-text with structured and symbolic representations, and to replace semantic evaluation with structure-based evaluation, thereby enabling low-cost and fine-grained control over the reasoning process when computing outcome rewards.
Notably, increasing reliance on process-based rewards may encourage models to over-optimize intermediate reasoning quality at the expense of correct final answers, giving rise to reward hacking \cite{Denison2024SycophancyTS}. This issue fundamentally reflects the limitations of conventional advantage estimation methods.

In this paper, we propose a \textbf{G}raph \textbf{R}easoning \textbf{P}aradigm (\textbf{GRP}), which realizes the structured and symbolic reasoning process using graph-structured representations.
As shown in Figure~\ref{fig:example}, the reasoning process in this paradigm is explicitly organized into a structured form, where each reasoning step is annotated with a specific cognitive label.
These labels correspond to different cognitive operations, including \emph{known}, \emph{generate}, \emph{aggregate}, \emph{reflect}, \emph{refine}, \emph{reverse}, and \emph{associate}. 
This structured representation transforms reasoning from unstructured text into an explicit graph, improving interpretability, and enabling systematic evaluation of reasoning processes. 
Based on this paradigm, we construct over 40.3k graph-structured mathematical reasoning chains and 12k graph-structured code generation chains with step-level labels, and perform supervised fine-tuning (SFT) to train LLMs to internalize graph-structured reasoning.

Building on the structured and symbolic outputs, we propose \textbf{PASC-GRPO}, a \textbf{P}rocess-\textbf{A}ware \textbf{S}tratified \textbf{C}lipping extension of \textbf{G}roup \textbf{R}elative \textbf{P}olicy \textbf{O}ptimization to further exert finer control over the reasoning process. 
We design a set of graph-based outcome rewards that evaluate the quality of the reasoning process using structural properties such as label validity, reachability, connectivity, informative subgraph, and reverse search consistency. 
These rewards do not rely on value models for semantic evaluation, leading to better generalization and improved training efficiency. 
To mitigate reward hacking introduced by multiple rewards, we further propose stratified clipping advantage estimation, which separately normalizes rewards within correct and incorrect groups, ensuring stable and reliable optimization.


Experiments across mathematical reasoning and code generation benchmarks demonstrate that our method consistently outperforms strong baselines, particularly on competition-level tasks. Furthermore, it effectively mitigates reward hacking and reduces reasoning length, improving both accuracy and inference efficiency. In summary, our contributions are threefold:
\begin{itemize}
    \item We propose a \textbf{Graph Reasoning Paradigm} that enables structured and symbolic reasoning processes. We construct over \textbf{52k} graph-structured reasoning chains with step-level cognitive labels for SFT.
    \item We propose \textbf{PASC-GRPO}, a reinforcement learning method that leverages graph-structured outcome rewards and stratified clipping advantage estimation to improve reasoning quality while mitigating reward hacking.
    \item We empirically demonstrate that our approach significantly improves the reasoning performance of LLMs, providing evidence for the effectiveness of graph reasoning paradigm.
\end{itemize}

\section{Related Work}
\paragraph{Structured cognition.} Structured reasoning frameworks organize the reasoning process into explicit and manageable representations. Examples include Tree-of-Thoughts~\citep{Yao2023TreeOT}, Graph-of-Thoughts~\citep{Besta2023GraphOT}, and cognitive-architecture-inspired models~\citep{Sumers2023CognitiveAF}. However, these approaches primarily focus on prompting strategies, and rely on semantic evaluation or manual annotation.

\paragraph{RL-enhanced Reasoning.} Reinforcement learning has been widely applied to improve the reasoning capabilities of LLMs \cite{Xu2025TowardsLR,Zhang2025ASO,Yuan2024FreePR}. However, these methods often rely on coarse reward signals, making fine-grained process control and evaluation difficult, and increasing the risk of reward hacking~\citep{Denison2024SycophancyTS}. Detailed technical background is provided in Appendix~\ref{app:technical_background}.

\section{Methodology}
\begin{figure*}[t]
    \centering
    \includegraphics[width=\linewidth]{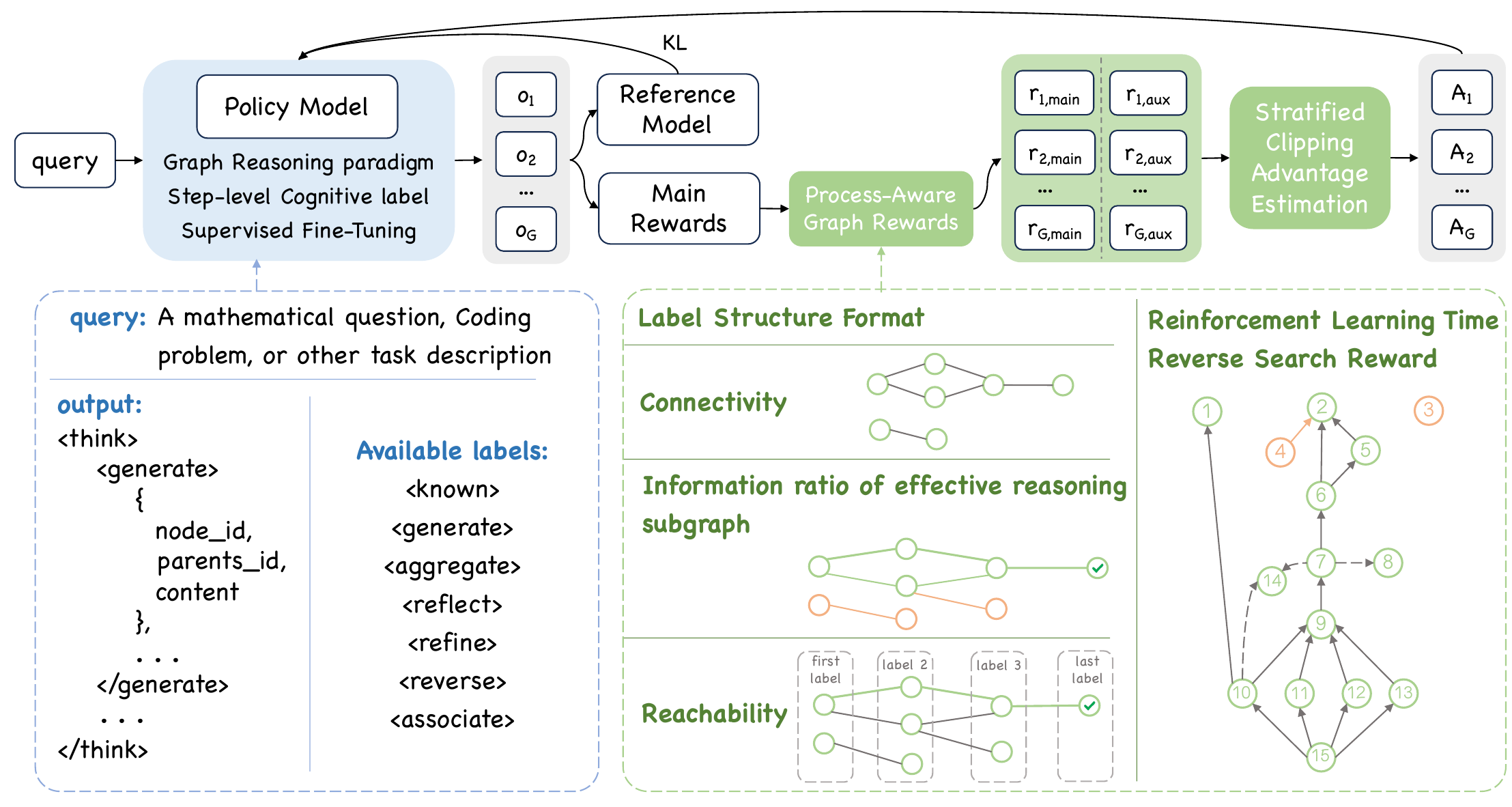}
    \caption{
    Overview of the proposed method.
    }
    \label{fig:overview}
\end{figure*}

As illustrated in Figure~\ref{fig:overview}, our method consists of two main components:
\begin{itemize}
    \item \textbf{Graph Reasoning Paradigm}, which enables LLMs to produce symbolized, interpretable, verifiable, and accurate graph-structured reasoning through supervised fine-tuning; 
    \item \textbf{PASC-GRPO}, a reinforcement learning framework that incorporates process-aware graph rewards and stratified clipping advantage estimation to further improve reasoning quality and training stability.
\end{itemize}

\subsection{Graph Reasoning Paradigm}
\label{sec:grp}

We design a graph-structured reasoning paradigm in which the model directly outputs its reasoning process as a graph.
This paradigm is realized through step-level cognitive labels, node-based reasoning representation, and data synthesis followed by supervised fine-tuning (SFT).

\subsubsection{Step-level Cognitive Labels}
\label{sec:labels}

Existing large reasoning models employ only coarse-grained \texttt{<think>} and \texttt{<answer>} tags, without further decomposition of the reasoning process.
As a result, emergent operations such as reflection may occur at arbitrary stages and with inconsistent content during reasoning.
To address this limitation, we introduce step-level reasoning tags.
Details are provided in Appendix~\ref{app:labels_description}.

By explicitly selecting a cognitive label at each step, the model is encouraged to reason about \emph{how} to emerge, not only \emph{can} emerge, which improves the interpretability of the reasoning process and the reliability of utilizing different reasoning patterns.

\subsubsection{Node-based Reasoning Representation}
\label{sec:nodes}

Under the graph reasoning paradigm, each cognitive label wraps one or more reasoning nodes.
Each node is defined as a tuple:
\begin{equation}
v = (\textit{id}, \textit{parents\_id}, \textit{content}),
\end{equation}
where \textit{id} is the unique identifier of the node, \textit{parents\_id} is a list of parent node ids, and \textit{content} denotes the content of the reasoning step.

This node-based representation enables the reasoning process to be abstracted into a directed graph, which forms the foundation for graph-structured, process-aware optimization.

\subsubsection{Data Synthesis and SFT}
\label{sec:data_sft}


To enhance the model's structural reasoning while maintaining its inherent inference fluidity, we implement an iterative synthesis pipeline. This process consists of three primary stages:

\paragraph{Generating and Validating Reasoning Traces.}
For each problem $p$ in the dataset $\mathcal{P}$, we first sample a reasoning trace using a teacher LLM $\mathcal{T}$ and verify its final answer against the ground truth.
If the answer is correct, the trace is retained for subsequent graph-structured transformation.
If the answer is incorrect, we trigger a regeneration process with an upper bound on the number of retries.
This regeneration continues until either a correct reasoning trace is obtained or the maximum retry limit is reached.
Only reasoning traces with verified correct answers are passed to the next stage.

\paragraph{Graph-Structured Translation.}
Given a validated reasoning trace $T$, we translate it into the Graph Reasoning Paradigm. This translation explicitly exposes the latent reasoning structure while preserving the original semantic content of the trace.
Graph representations derived from originally correct and regenerated traces are jointly collected for quality inspection. Detailed prompts and examples are provided in Appendix~\ref{app:translation}.

\paragraph{Graphical CoT Verification and Refinement.}
To ensure the quality of the synthesized graph-structured reasoning data, we introduce an automated quality control loop that evaluates each generated graph from two complementary aspects:

\begin{itemize}
    \item \textbf{Node--Label Consistency.}
    We verify that the reasoning content of each node conforms to the semantic scope of its assigned label.

    \item \textbf{Parent--Child Coherence.}
    We check whether each node is logically coherent with its parent node(s), ensuring a consistent and non-contradictory reasoning flow.
\end{itemize}

If a graph $G$ fails either of the above checks, we generate structured feedback that explicitly identifies the problematic nodes and the corresponding violation types.
This feedback, together with a predefined refinement prompt template, is fed back to the LLM to trigger a re-translation step.
The refinement process iterates until the graph passes all quality checks or a maximum number of translation attempts is reached. 
Detailed prompts and cases are provided in Appendix~\ref{app:verify}.

\paragraph{Supervised Fine-Tuning with Mixed Reasoning Formats.}  
After constructing the graph-structured reasoning dataset, we perform supervised fine-tuning on both graph-structured traces and a subset of original CoT traces that could not be converted into valid graphs. Although these samples do not conform to the graph format, their problem–solution pairs and coherent reasoning steps remain informative. This mixed-format training allows the model to internalize the graph reasoning paradigm while preserving fluency, completeness, and generalization of its native reasoning behavior.

\subsection{PASC-GRPO}
\label{sec:pasc_grpo}

Fine-grained control over the reasoning process is crucial for solving complex reasoning tasks. \cite{Shao2025DeepSeekMathV2TS}
After SFT, the model internalizes the graph reasoning paradigm, where the reasoning process can be abstracted as a graph independent of specific reasoning content.
Building on it, we propose \textbf{PASC-GRPO} (\textbf{P}rocess-\textbf{A}ware \textbf{S}tratified \textbf{C}lipping GRPO), which enables optimization of reasoning length and process quality.

\subsubsection{Process-Aware Graph Rewards}
\label{sec:graph_rewards}

We use NetworkX~\cite{SciPyProceedings_11} to construct reasoning graphs.
On this basis, we design \emph{Process-Aware Graph Rewards}, which can be grouped into three categories:

\begin{itemize}
    \item \textbf{Graph structure format rewards}, including the \emph{Label Structure Format Reward}, which enforce valid and well-formed reasoning graphs.
    \item \textbf{Reasoning length rewards}, including the \emph{Connectivity Reward}, which encourages fewer connected components, and the \emph{Information Ratio of Effective Reasoning Subgraph}, which promotes shorter reasoning paths within each component.
    \item \textbf{Reasoning process quality rewards}, including the \emph{Reachability Reward} at the global process level, and the \emph{Reinforcement Learning Time Reverse Search Reward} at the step level.
\end{itemize}

\paragraph{Label Structure Format Rewards.}
We introduce it to enforce the structural validity of graph-structured reasoning.
Let $\mathcal{G}=(\mathcal{V}, \mathcal{E})$ denote the reasoning graph, where $Pa(v)$ is the set of parent nodes of $v$.
Let $\mathcal{K}$, $\mathcal{A}$, and $\mathcal{R}$ denote \textit{known}, \textit{aggregate}, and \textit{refine} nodes, and $\mathcal{T}$ denote the set of reasoning tags. The overall format reward $R_{fmt}$ is defined as the average of three complementary sub-rewards:
\begin{equation}
R_{fmt}
=
\frac{1}{3}
\left(
R_{dens}
+
R_{topo}
+
R_{para}
\right).
\end{equation}

\begin{itemize}

\item \textbf{Node Density.}
For \textit{aggregate} and \textit{refine}, each tag is required to wrap exactly one node:
\begin{equation}
R_{dens}
=
\frac{1}{|\mathcal{A} \cup \mathcal{R}|}
\sum_{v \in \mathcal{A} \cup \mathcal{R}}
\mathds{1}
\big(
|v|_{\text{label}} = 1
\big).
\end{equation}

\item \textbf{Topological Validity.} Specifically, \textit{known} nodes must not depend on prior reasoning, \textit{aggregate} nodes must combine multiple predecessors, and \textit{refine} nodes must extend exactly one preceding step:
\begin{equation}
R_{topo}
=
\frac{1}{|\mathcal{K} \cup \mathcal{A} \cup \mathcal{R}|}
\sum_{v \in \mathcal{K} \cup \mathcal{A} \cup \mathcal{R}}
\Phi(v),
\end{equation}
\begin{equation}
\Phi(v)
=
\begin{cases}
1, & |Pa(v)| = 0,\ v \in \mathcal{K}, \\
1, & |Pa(v)| > 1,\ v \in \mathcal{A}, \\
1, & |Pa(v)| = 1,\ v \in \mathcal{R}, \\
0, & \text{otherwise}.
\end{cases}
\end{equation}

\item \textbf{Parallelism.}
Nodes within a tag are treated as parallel reasoning units and are therefore prohibited from forming parent--child relations:
\begin{equation}
\begin{aligned}
R_{para}
&=
\frac{1}{|\mathcal{T}|}
\sum_{T \in \mathcal{T}}
\mathds{1}
\Big(
\forall\, v_i, v_j \in T, \\
&\qquad
v_i \notin Pa(v_j)
\Big)
\end{aligned}
\end{equation}
\end{itemize}

\paragraph{Connectivity Reward.} To discourage fragmented reasoning, we define the connectivity reward based on the number of connected subgraphs:
\begin{equation}
R_{conn}
=
\frac{1}{n},
\end{equation}
where $n$ is the number of connected subgraphs in $\mathcal{G}$. This reward favors reasoning with fewer isolated branches, resembling more focused reasoning.

\paragraph{Information Ratio of Effective Reasoning Subgraph.}
We define the effective reasoning subgraph (ERS) as the part of the reasoning graph that effectively contributes to reaching the final answer.
The shortest path from the initial conditions to the answer node forms a backbone of effective reasoning.
Reasoning branches that diverge from this backbone but eventually merge back are also considered effective.
Branches that do not reconnect are treated as ineffective explorations, as they may correspond to incorrect reasoning attempts.
\begin{equation}
\mathcal{V}_{ERS}
=
\{\, v \in \mathcal{V} \mid v_{start} \rightsquigarrow v \rightsquigarrow v_{end} \,\}.
\end{equation}

The ERS Information Ratio $R_{ers}$ is measured by token count $I(\cdot)$:
\begin{equation}
    R_{ent} = \frac{\sum_{v \in \mathcal{V}_{ERS}} I(v)}{\sum_{v \in \mathcal{V}_{Total}} I(v)}
\end{equation}
This encourages the model to avoid "dead-end" branches and redundant chatter.

\begin{figure*}[t]
    \centering
    \includegraphics[width=\linewidth]{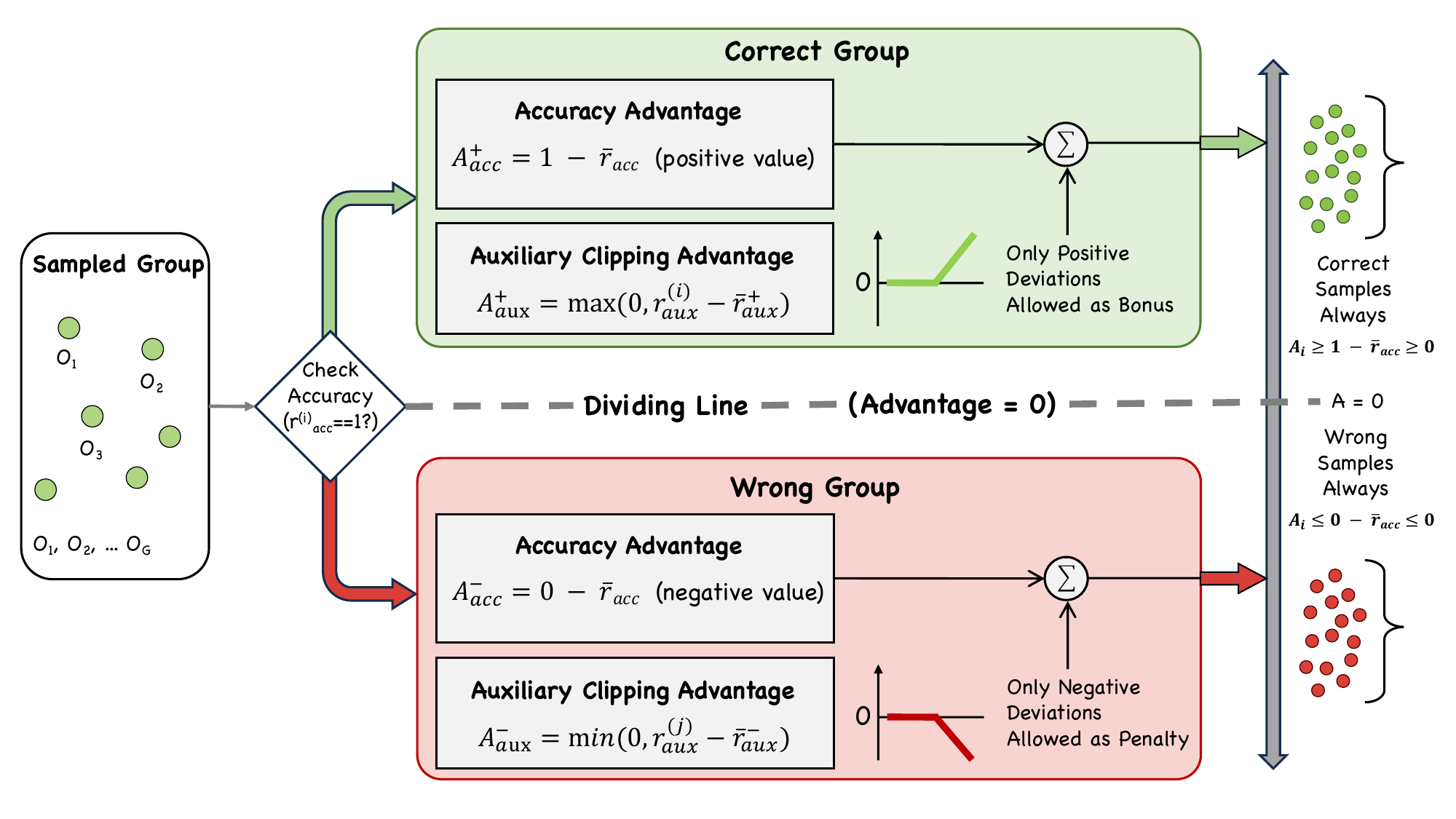}
    \caption{
    Illustration of the Stratified Clipping Advantage Estimation.
    }
    \label{fig:scae}
\end{figure*}

\paragraph{Reachability Reward.}
To mitigate the "wrong process, correct answer" phenomenon \cite{Akter2025InducingFI}, we evaluate the end-to-end logical flow:
\begin{equation}
R_{reach} = \mathds{1}(v_{start} \rightsquigarrow v_{end})
\end{equation}
where $R_{reach} = 1$ if there exists a directed path in $\mathcal{G}$, ensuring the reasoning process is complete and leads to the final answer.

\paragraph{Reinforcement Learning Time Reverse Search Reward.}
Unlike forward heuristics such as MCTS or PRM, which incrementally expand reasoning paths to approximate the correct answer, our reward directly leverages the known answer. 
After the reasoning graph $\mathcal{G}$ is completed, we traverse it backward from the answer node $v_{ans}$ to identify which nodes actually contribute to reaching the answer.For each node $v \in \mathcal{V}$, we assign
\begin{equation}
r(v) =
\begin{cases}
1 & \text{if } v \rightsquigarrow v_{ans}, \\
0 & \text{otherwise,}
\end{cases}
\end{equation}
where $v \rightsquigarrow v_{ans}$ indicates that $v$ is reachable from $v_{ans}$ via backward traversal.
The total reward is computed as the average over all nodes:
\begin{equation}
R_{rev} = \frac{1}{|\mathcal{V}|} \sum_{v \in \mathcal{V}} r(v).
\end{equation}

\paragraph{Overall Process-Aware Graph Reward.}
The final process-aware graph reward is defined as a weighted sum with scalar weights $w_1,\dots,w_5$, where each component reward is normalized to $[0,1]$ and $\sum_{i=1}^{5} w_i = 1$:

\begin{equation}
\begin{aligned}
R_{\text{graph}}
&=
w_1 R_{fmt}
+ w_2 R_{conn}
+ w_3 R_{ers} \\
&\quad
+ w_4 R_{reach}
+ w_5 R_{rev}
\end{aligned}
\end{equation}

\subsubsection{Stratified Clipping Advantage Estimation}

To mitigate the reward hacking phenomenon — where the model might optimize graph-structured rewards despite producing incorrect answers — we propose the \textbf{Stratified Clipping Advantage Estimation} (SCAE) method. This approach hierarchically prioritizes task accuracy over auxiliary structural signals. The logical flow of the proposed SCAE is illustrated in Figure \ref{fig:scae}.

\paragraph{Group Stratification.}
For a sampled group of $G$ reasoning traces, we first calculate the mean accuracy reward $\bar{r}_{acc} \in [0, 1]$. The group is then partitioned into a \textbf{Correct Group} ($\mathcal{G}_{corr}$) where $r_{acc}^{(i)}=1$ and a \textbf{Wrong Group} ($\mathcal{G}_{wrong}$) where $r_{acc}^{(j)}=0$. We define the baseline accuracy advantages for these strata as:
\begin{equation}
    A_{acc} = 
    \begin{cases} 
    1 - \bar{r}_{acc}, & \text{if } i \in \mathcal{G}_{corr} \\
    0 - \bar{r}_{acc}, & \text{if } j \in \mathcal{G}_{wrong}
    \end{cases}
\end{equation}
This ensures the baseline advantage is non-negative ($A_{acc} \ge 0$) for the correct group and non-positive ($A_{acc} \le 0$) for the wrong group.

\paragraph{Asymmetrical Auxiliary Clipping.}  
Within each stratum, we calculate the mean of the auxiliary graph rewards, denoted as $\bar{r}_{aux}^+$ for the correct group and $\bar{r}_{aux}^-$ for the wrong group. Here, the auxiliary reward refers to all rewards other than the main reward (e.g. accuracy reward), such as graph or format rewards. To prevent structural rewards from overriding the accuracy signal, we apply asymmetrical clipping:

\begin{itemize}

    \item \textbf{Correct Group Clipping:} In $\mathcal{G}_{corr}$, structural rewards are treated strictly as bonuses. Even if a trace's structural reward $r_{aux}^{(i)}$ is below the group mean, it is not penalized, ensuring the final advantage $A_i$ never falls below the accuracy baseline:
    \begin{equation}
        A_i = (1 - \bar{r}_{acc}) + \max(0, r_{aux}^{(i)} - \bar{r}_{aux}^+)
    \end{equation}
    to ensure $i \in \mathcal{G}_{corr}$, $A_i \ge 1 - \bar{r}_{acc} \ge 0$.

    \item \textbf{Wrong Group Clipping:} In $\mathcal{G}_{wrong}$, structural rewards serve exclusively as penalties. No matter how high the structural quality $r_{aux}^{(j)}$ is relative to the mean, it receives no credit, ensuring the advantage remains non-positive:
    \begin{equation}
        A_j = (0 - \bar{r}_{acc}) + \min(0, r_{aux}^{(j)} - \bar{r}_{aux}^-)
    \end{equation}
    to ensure $j \in \mathcal{G}_{wrong}$, $A_j \le 0 - \bar{r}_{acc} \le 0$.
\end{itemize}

In summary, SCAE offers the following advantages for training graph-reasoning models:

\begin{table*}[htbp]
    \centering
    \caption{Experimental results on Mathematical Reasoning and Code Generation benchmarks.}
    \label{tab:main_results}
    \small
    \renewcommand{\arraystretch}{1.2} 
    \begin{tabular}{lccccc}
        \toprule
        \rowcolor{gray!15}
        \multicolumn{6}{c}{\textbf{Mathematical Reasoning}} \\
        \midrule
        \textbf{Model} & \textbf{GSM8K} & \textbf{MATH500} & \textbf{AMC23} & \textbf{AIME24} & \textbf{AIME25} \\
        \midrule
        LLaMA-3.1-8B-Instruct & 85.00 & 54.80 & 41.20 & 6.30 & 2.70  \\
        Gemma-3-12B-IT         & 94.40 & 85.60 & 77.30 & 22.40 & 18.80  \\
        Gemma-3-27B-IT         & \textbf{95.90} & 90.00 & 80.50 & 32.60 & 24.00  \\
        \midrule
        Qwen3-4B-Base               & 70.20 & 55.43 & 19.17 & 10.00 & 6.33  \\
        \rowcolor{cyan!5} 
        ~+ GRP-SFT             & 82.30 & 72.12 & 63.33 & 26.67 & 18.70  \\
        \rowcolor{cyan!5} 
        ~+ PASC-GRPO           & 90.45 & 85.32 & 71.43 & 33.90 & 24.13  \\
        \midrule
        Qwen3-8B-Base               & 73.01 & 60.08 & 38.12 & 10.00 & 7.60  \\
        \rowcolor{cyan!5} 
        ~+ GRP-SFT             & 87.72 & 81.60 & 75.00 & 40.00 & 33.33  \\
        \rowcolor{cyan!5} 
        ~+ PASC-GRPO           & 95.37 & \textbf{91.20} & \textbf{82.50} & \textbf{46.67} & \textbf{38.79}  \\
        \midrule 
        \rowcolor{gray!15}
        \multicolumn{6}{c}{\textbf{Code Generation}} \\
        \midrule
        \textbf{Model} & \textbf{MBPP} & \textbf{MBPP+} & \textbf{HumanEval} & \textbf{HumanEval+} & \textbf{LiveCodeBench} \\
        \midrule
        LLaMA-3.1-8B-Instruct & 61.20 & 52.30 & 69.70 & 62.80 & 10.80  \\
        Gemma-3-12B-IT         & 73.00 & 62.10 & 85.40 & 78.20 & 25.70  \\
        Gemma-3-27B-IT         & 74.40 & 63.00 & 87.80 & 80.00 & 26.90  \\
        \midrule
        Qwen3-4B-Base               & 62.40 & 51.30 & 75.60 & 70.70 & 21.14  \\
        \rowcolor{cyan!5} 
        ~+ GRP-SFT             & 65.43 & 54.2 & 81.67 & 76.90 & 42.93  \\
        \rowcolor{cyan!5} 
        ~+ PASC-GRPO           & 67.33 & 55.43 & 82.45 & 77.83 & 46.79  \\
        \midrule
        Qwen3-8B-Base               & 73.27 & 61.73 & 80.47 & 74.27 & 29.76  \\
        \rowcolor{cyan!5} 
        ~+ GRP-SFT             & 75.45 & 64.71 & 86.79 & 82.09 & 53.93  \\
        \rowcolor{cyan!5} 
        ~+ PASC-GRPO           & \textbf{77.49} & \textbf{65.24} & \textbf{88.43} & \textbf{83.98} & \textbf{56.12}  \\
        \bottomrule
    \end{tabular}
\end{table*}

\begin{itemize}
    \item \textbf{Accuracy Primacy:} By setting a hard floor for correct samples ($A_i \ge A_{acc}^+ \ge 0$) and a hard ceiling for incorrect samples ($A_j \le A_{acc}^- \le 0$), SCAE ensures that correctness remains the dominant optimization objective.
    \item \textbf{Reward Hacking Resilience:} The asymmetric clipping mechanism prevents the model from "cheating" by generating high-quality graph structures for incorrect derivations to offset the accuracy penalty.
\end{itemize}

\section{Experiments}

\subsection{Experimental Setup}

\paragraph{Baselines.} We include LLaMA-3.1-8B-Instruct from Meta \cite{Dubey2024TheL3} and Gemma-3-12B-IT / Gemma-3-27B-IT from Google \cite{Kamath2025Gemma3T} as comparison baselines.  
To evaluate the effectiveness of GRP and PASC-GRPO, we train on the pre-trained Qwen3-4B-Base and Qwen3-8B-Base models from Qwen \cite{Yang2025Qwen3TR}. Detailed training data construction procedures are provided in Appendix~\ref{app:training_data}.

\paragraph{Benchmarks.} For mathematical reasoning, the benchmarks includes GSM8K \cite{Cobbe2021TrainingVT}, MATH500 \cite{Lightman2023LetsVS}, and competition-level benchmarks such as AMC23 \cite{amc23_dataset}, AIME24 \cite{aime24_dataset}, and AIME25 \cite{aime25_dataset}. For code generation, we utilize the widely-used MBPP \cite{Austin2021ProgramSW}, MBPP+ \cite{Liu2023IsYC}, HumanEval \cite{Chen2021EvaluatingLL}, HumanEval+ \cite{Liu2023IsYC}, and the highly challenging LiveCodeBench v5 \cite{Jain2024LiveCodeBenchHA}.

\paragraph{Evaluation Metrics.} For larger datasets (GSM8K, MATH500, and LiveCodeBench), we report the avg@3 accuracy; for other benchmarks, we report the avg@16 accuracy. Considering the extensive derivation required for competition-level mathematics (AIME24, AIME25), we set the \texttt{max\_completion\_tokens} to 32K, while maintaining a limit of 8K for other datasets.


\subsection{Main Results}

\paragraph{Performance Overview.} 
Table~\ref{tab:main_results} reports the performance of our models on mathematical reasoning and code generation benchmarks. 
Both GRP-SFT and PASC-GRPO consistently improve accuracy, demonstrating that the proposed GRP effectively enhances reasoning capabilities.

\paragraph{Impact of GRP-SFT.}
GRP-SFT brings substantial gains in mathematical reasoning. For Qwen3-8B-Base, accuracy improves by 21.52\% on MATH500 and 18.30\% on GSM8K, with a particularly large gain of 36.88\% on the competition-level AMC23 benchmark. Code generation tasks also benefit from GRP-SFT, indicating that the proposed graph-based reasoning paradigm generalizes beyond mathematics.

\paragraph{Effect of PASC-GRPO.}
PASC-GRPO further enhances performance, especially on high-difficulty benchmarks. On AIME24, Qwen3-4B and Qwen3-8B achieve additional gains of 7.23\% and 6.67\% over the SFT stage, respectively. These results suggest that the process-aware reward are effective for multi-step problems.

\paragraph{Comparison with Open-Source LLMs.}
Despite using fewer parameters and training resources, our best model surpasses widely adopted open-source models on competition-level benchmarks.  In particular, it surpasses Gemma-3-27B-IT by over 14\% accuracy on both AIME24 and AIME25, and achieves comparable or better performance on standard math and code benchmarks such as MATH500, AMC23, MBPP+, and HumanEval+. These results highlight the effectiveness of GRP and PASC-GRPO.

\subsection{Ablation Studies}

\begin{table}[htbp]
\centering
\small
\setlength{\tabcolsep}{4pt} 
\caption{Ablation Experiment Results. Accuracy (\%).}
\label{tab:ablation_combined}
\begin{tabular}{lccc} 
\toprule
\textbf{Method} & \textbf{MATH500} & \textbf{AMC23} & \textbf{AIME24} \\
\midrule
Qwen3-8B-Base & 60.08 & 38.12 & 10.00 \\
\midrule
GRP-SFT & 81.60 & 75.00 & 40.00 \\
SFT & 76.45 & 67.80 & 31.50 \\
\midrule
PASC-GRPO & 91.20 & 82.50 & 46.67 \\
GRPO & 88.75 & 79.85 & 43.80 \\
\midrule
w/o Reachability & 87.45 & 79.30 & 40.15 \\
w/o Reach. \& RSR & 83.12 & 76.55 & 35.80 \\
\midrule
w/o SCAE & 84.15 & 76.20 & 32.50 \\
\bottomrule
\end{tabular}
\end{table}

\paragraph{Ablation of GRP and PASC-GRPO.}

We study the effectiveness of GRP and PASC-GRPO.
For SFT, we use the same data as GRP-SFT but replace graph-structured reasoning with standard chain-of-thought.
As shown in Table~\ref{tab:ablation_combined}, this variant performs worse than GRP-SFT on all benchmarks. 

For RL, we compare PASC-GRPO with the original GRPO under identical training settings.
GRPO consistently yields lower performance.
These results show that both graph-structured reasoning and process-aware rewards are essential for strong reasoning performance.

\paragraph{Ablation of Reasoning-Length Rewards.}

\begin{figure}[t]
    \centering
    \includegraphics[width=\columnwidth]{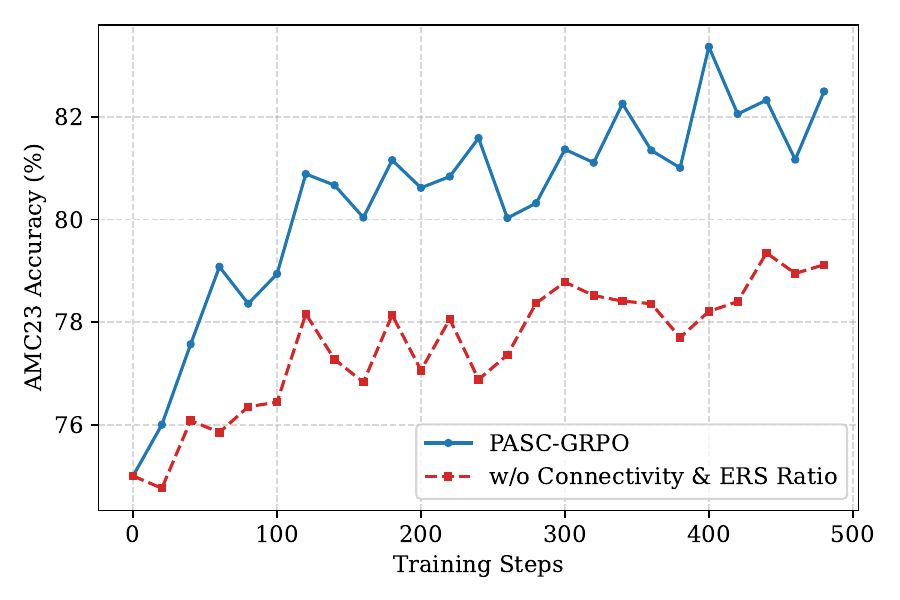}
    \vspace{0.6em}
    \includegraphics[width=\columnwidth]{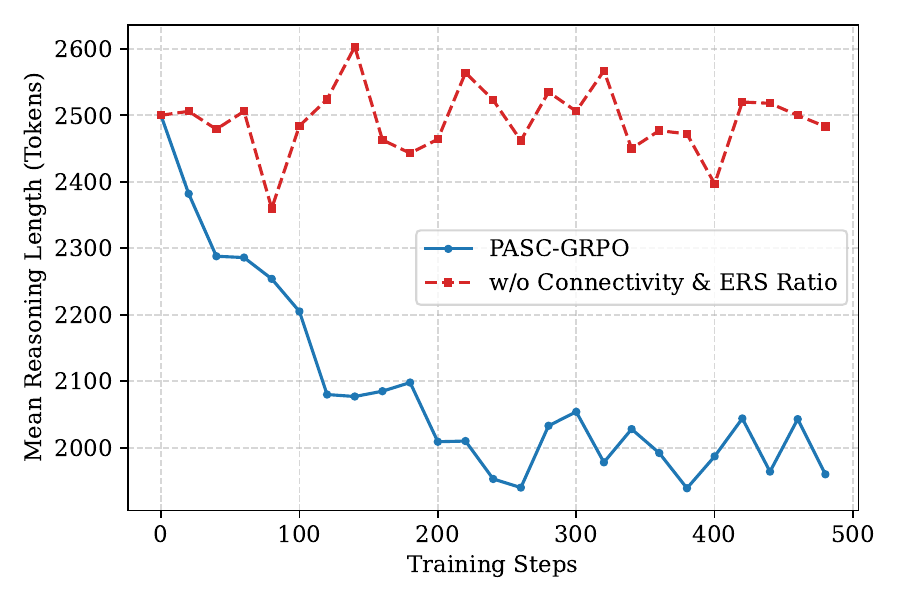}
    \caption{
    (Top) Accuracy comparison between PASC-GRPO and its variants without connectivity and ERS Ratio rewards on AMC23.
    (Bottom) Corresponding changes in mean reasoning length during training.
    }
    \label{fig:amc23_training_dynamics}
\end{figure}

We remove the connectivity reward and further remove ERS Ratio, starting from the GRP-SFT baseline.
Figure~\ref{fig:amc23_training_dynamics} shows the changes in accuracy and length during training, comparing the cases with and without connectivity and ERS Ratio rewards. For more details on the ablation studies, see Appendix~\ref{app:ablation}.

\paragraph{Ablation of Reasoning process quality rewards.}

Table~\ref{tab:ablation_combined} reports the ablation results of the reachability and reverse search rewards, starting from the GRP-SFT baseline.
Removing the reachability reward leads to a performance drop across all benchmarks.
Further removing the reverse search reward causes a more substantial degradation.
These results indicate that both rewards are critical for guiding effective graph reasoning.

\paragraph{Ablation of SCAE.}

Table~\ref{tab:ablation_combined} shows that using the original GRPO instead of Stratified Clipping Advantage Estimation (SCAE) consistently degrades performance across all benchmarks.
These results suggest that SCAE is crucial for stabilizing the reasoning process and improving accuracy.

\subsection{Discovery}

As shown in Table~\ref{tab:find}, our best model trained from Qwen3-8B-Base matches or surpasses the performance of Qwen3-8B on hard code generation benchmarks. 
Qwen3-8B is a state-of-the-art model distilled using extensive human and computational resources.  
These results highlight the effectiveness of our methods in handling complex reasoning tasks, especially code generation.

\begin{table}[htbp]
\centering
\small
\caption{Performance Comparison on Code Generation}
\label{tab:find}
\begin{tabular}{lccc}
\toprule
\textbf{Model} & \textbf{MBPP} & \textbf{MBPP+} & \textbf{LiveCodeBech}\\
\midrule
qwen3-8B & 76.70 & 65.10 & 57.29 \\
\midrule
qwen3-8B-Base & 73.27 & 61.73 & 29.76 \\
~+ GRP-SFT  & 75.45 & 64.71 & 53.93 \\
~+ PASC-GRPO  & 77.49 & 65.24 & 56.12 \\
\bottomrule
\end{tabular}
\end{table}

\section{Conclusion}

In this paper, we propose the Graph Reasoning Paradigm (GRP), which shifts LLM reasoning from plain text into structured, symbolic graph representations. We further introduce PASC-GRPO, a reinforcement learning framework that leverages topology-aware rewards and Stratified Clipping Advantage Estimation to enhance reasoning quality while mitigating reward hacking. Our experiments across math and coding demonstrate that GRP significantly improves performance. This work underscores the potential of structured symbolic reasoning and topology-aware rewards in developing more reliable and efficient reasoning models.

\section*{Limitations}

While the Graph Reasoning Paradigm and PASC-GRPO demonstrate substantial improvements in mathematical reasoning and code generation, several promising directions remain for future exploration. Currently, our evaluation primarily focuses on these logic-intensive domains; extending the graph-structured thinking framework to broader tasks, such as commonsense reasoning or agentic planning, presents a valuable opportunity to further assess its generalization capabilities. Additionally, while we have verified the effectiveness of our approach on 4B and 8B parameter models, investigating the scaling laws of graph reasoning on larger-scale foundation models could reveal further performance gains. Finally, future work may also explore more dynamic and adaptive graph topology generation methods to support increasingly complex and flexible cognitive processes without relying on predefined cognitive label sets.
 
\bibliography{references}

\newpage
\appendix

\section{Technical Background}
\label{app:technical_background}
\subsection{Group Relative Policy Optimization (GRPO)}

GRPO \cite{Shao2024DeepSeekMathPT} is a reinforcement learning algorithm designed for training language models with group-based advantage estimation. For each question $q$, a group of outputs $\{o_1, o_2, \cdots, o_G\}$ are sampled from the old policy model $\pi_{\theta_{old}}$. and optimizes the current policy $\pi_{\theta}$ using:

\begin{equation}
\begin{aligned}
\mathcal{J}_{\mathrm{GRPO}}(\theta)
=
\mathbb{E}_{q \sim P(Q),\, \{o_i\}_{i=1}^{G} \sim \pi_{\theta_{\mathrm{old}}}(\cdot \mid q)}
\\
\frac{1}{G}
\sum_{i=1}^{G}
\frac{1}{|o_i|}
\sum_{t=1}^{|o_i|}
\Big(
\min \Big[
r_{i,t}(\theta)\,\hat{A}_{i,t},\,
\\
\mathrm{clip}\!\left(
r_{i,t}(\theta),\, 1-\epsilon,\, 1+\epsilon
\right)\hat{A}_{i,t}
\Big]
\\
-
\beta\, \mathrm{D}_{\mathrm{KL}}
\big(
\pi_\theta \,\|\, \pi_{\mathrm{ref}}
\big)
\Big).
\end{aligned}
\end{equation}

\begin{equation}
r_{i,t}(\theta)
=
\frac{
\pi_\theta(o_{i,t} \mid q, o_{i,<t})
}{
\pi_{\theta_{\mathrm{old}}}(o_{i,t} \mid q, o_{i,<t})
}.
\end{equation}

\begin{equation}
\hat{A}_{i,t}
=
\tilde{r}_i
=
\frac{
r_i - \mathrm{mean}(r)
}{
\mathrm{std}(r)
}.
\end{equation}

\subsection{Outcome Rewards and Process Rewards}

In reinforcement learning for LLMs, rewards can be categorized into two types based on when they are assigned during the generation process.

\textbf{Outcome Rewards} evaluate the final result of a complete generation sequence. Given a state-action trajectory $(s_1, a_1, \ldots, s_T, a_T)$, the outcome reward is defined as $R = r(s_T)$, where only the terminal state $s_T$ receives a reward signal. This approach is commonly used in tasks where quality can only be judged after seeing the complete output, such as code correctness or final answer accuracy.

\textbf{Process Rewards}, on the contrary, provide feedback at each intermediate step of the reasoning process. The cumulative reward is computed as $R = \sum_{t=1}^{T} r(s_t, a_t)$, where $r(s, a)$ assigns credit to each reasoning step. Process rewards enable more fine-grained supervision and can guide the model toward correct reasoning paths even when the final answer is incorrect.

\subsection{Monte Carlo Tree Search (MCTS)}

Monte Carlo Tree Search \cite{Xie2024MonteCT} is a heuristic search algorithm that builds a search tree by iteratively selecting, expanding, simulating, and backpropagating rewards. In the context of LLM reasoning, each node represents a partial generation state $s$, and edges represent token or sequence actions $a$.

The algorithm balances exploration and exploitation using the Upper Confidence Bound (UCB) formula:

\begin{equation}
\mathrm{UCB}(s, a)
=
Q(s, a)
+
c \sqrt{
\frac{\ln N(s)}{N(s, a)}
}.
\end{equation}

where $Q(s, a)$ is the estimated value of taking action $a$ in state $s$, $N(s)$ is the visit count of state $s$, $N(s, a)$ is the visit count of the state-action pair, and $c$ is the exploration constant.

\section{Step-level Cognitive Labels}
\label{app:labels_description}

\begin{figure}[t]
    \centering
    \includegraphics[width=0.86\linewidth]{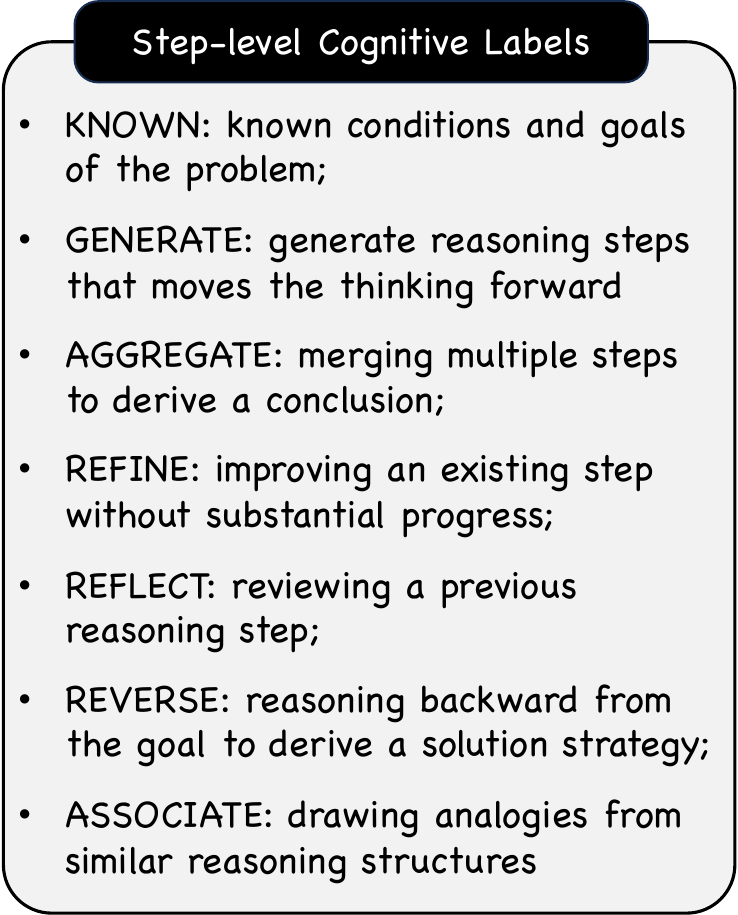}
    \caption{
    Step-level Cognitive Labels.
    }
    \label{fig:labels}
\end{figure}

Figure~\ref{fig:labels} illustrates the step-level cognitive labels used in our Graph Reasoning Paradigm. 
Each reasoning step in the structured graph is annotated with one of these labels, which represent distinct cognitive operations such as \emph{known}, \emph{generate}, \emph{aggregate}, \emph{reflect}, \emph{refine}, \emph{reverse}, and \emph{associate}. 
These labels enable fine-grained symbolic abstraction of the reasoning process, facilitating both interpretability and structured evaluation of model behavior.

\section{Graph-Structured Translation Prompt and Examples}
\label{app:translation}
\subsection{Graph-Structured Translation Prompt}
\begin{figure*}[t]
    \centering
    \includegraphics[width=\linewidth]{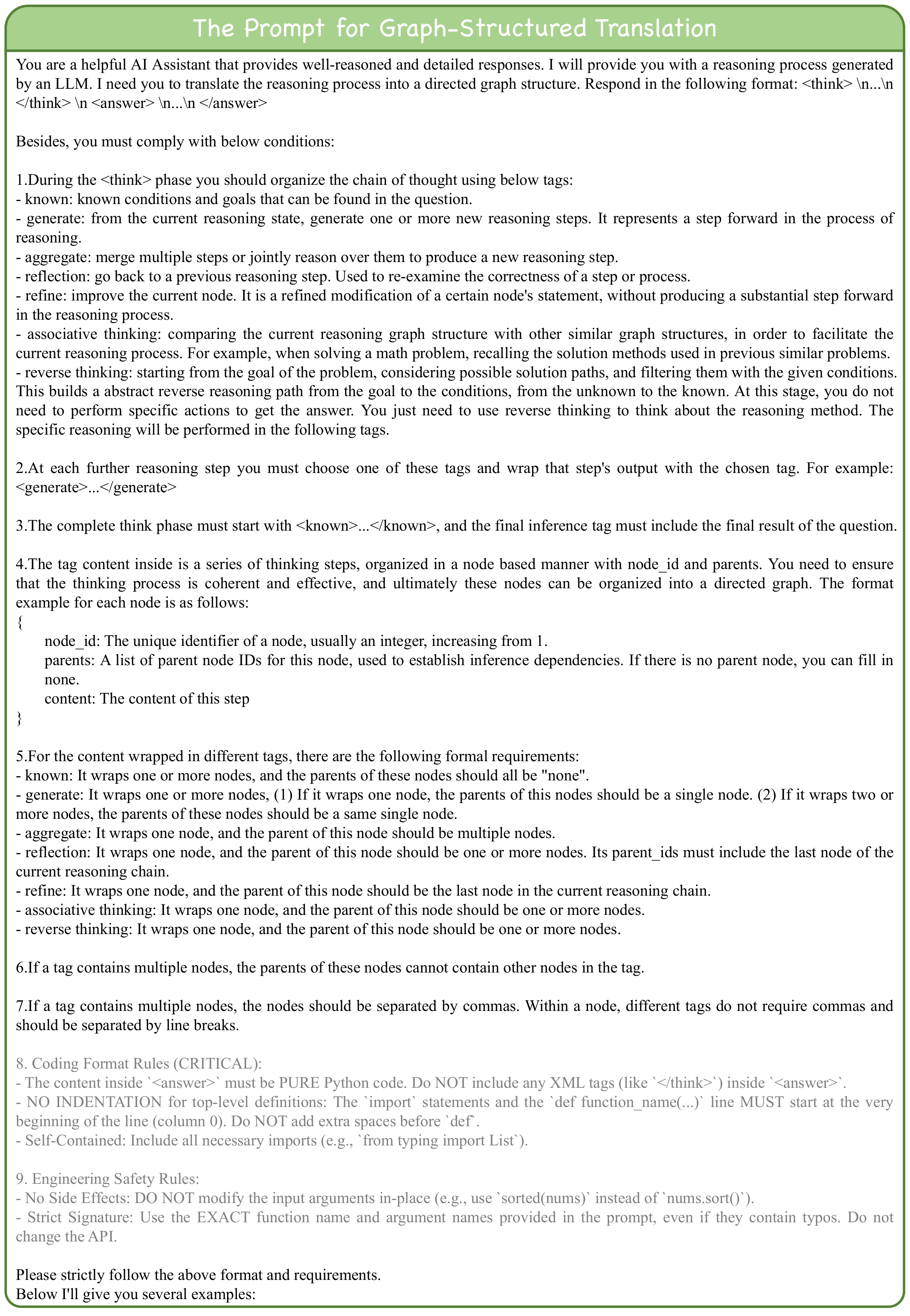}
    \caption{
    The prompt template for ``Graph-Structured Translation''. Criteria 8 and 9 are specifically added to accommodate code generation task.
    }
    \label{fig:prompt_grpah}
\end{figure*}

We design a unified prompt for graph-structured reasoning translation,
which specifies the node schema, thinking-mode tags, and dependency relations.
The prompt is shown in Figure~\ref{fig:prompt_grpah}.
The illustrative examples mentioned in the prompt are presented in the following subsections. The original Chain-of-Thought traces are generated using \texttt{Qwen3-Max-Preview}, which provides explicit thinking processes, and are subsequently translated into graph-structured representations using the \texttt{Qwen3-Max} model.

\subsection{Mathematical Examples}
\begin{figure*}[htbp]
    \centering
    \includegraphics[width=\textwidth]{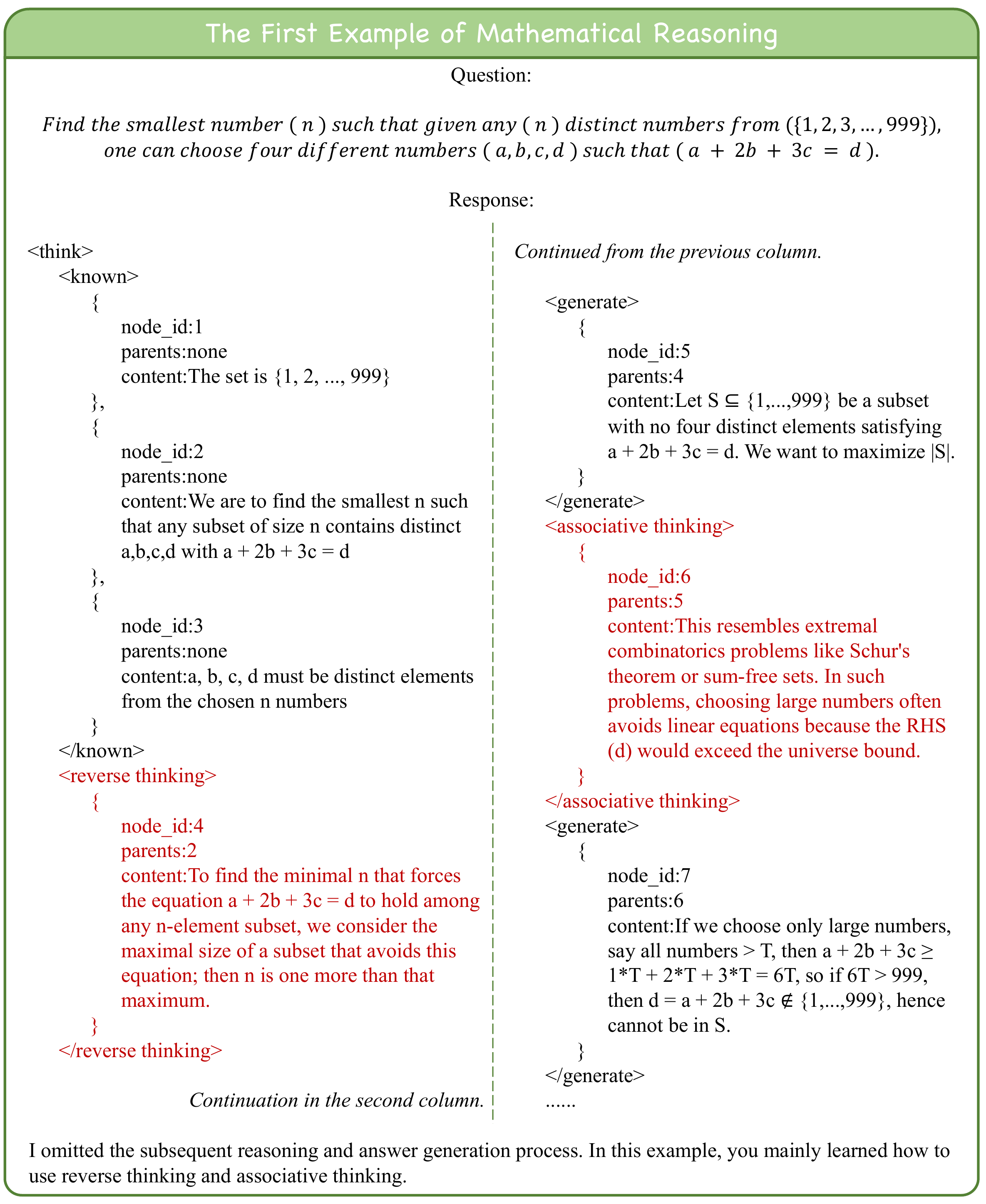}
    \caption{A mathematical example illustrating reverse and associative thinking.}
    \label{fig:math_example_reverse}
\end{figure*}

\begin{figure*}[htbp]
    \centering
    \includegraphics[width=\textwidth]{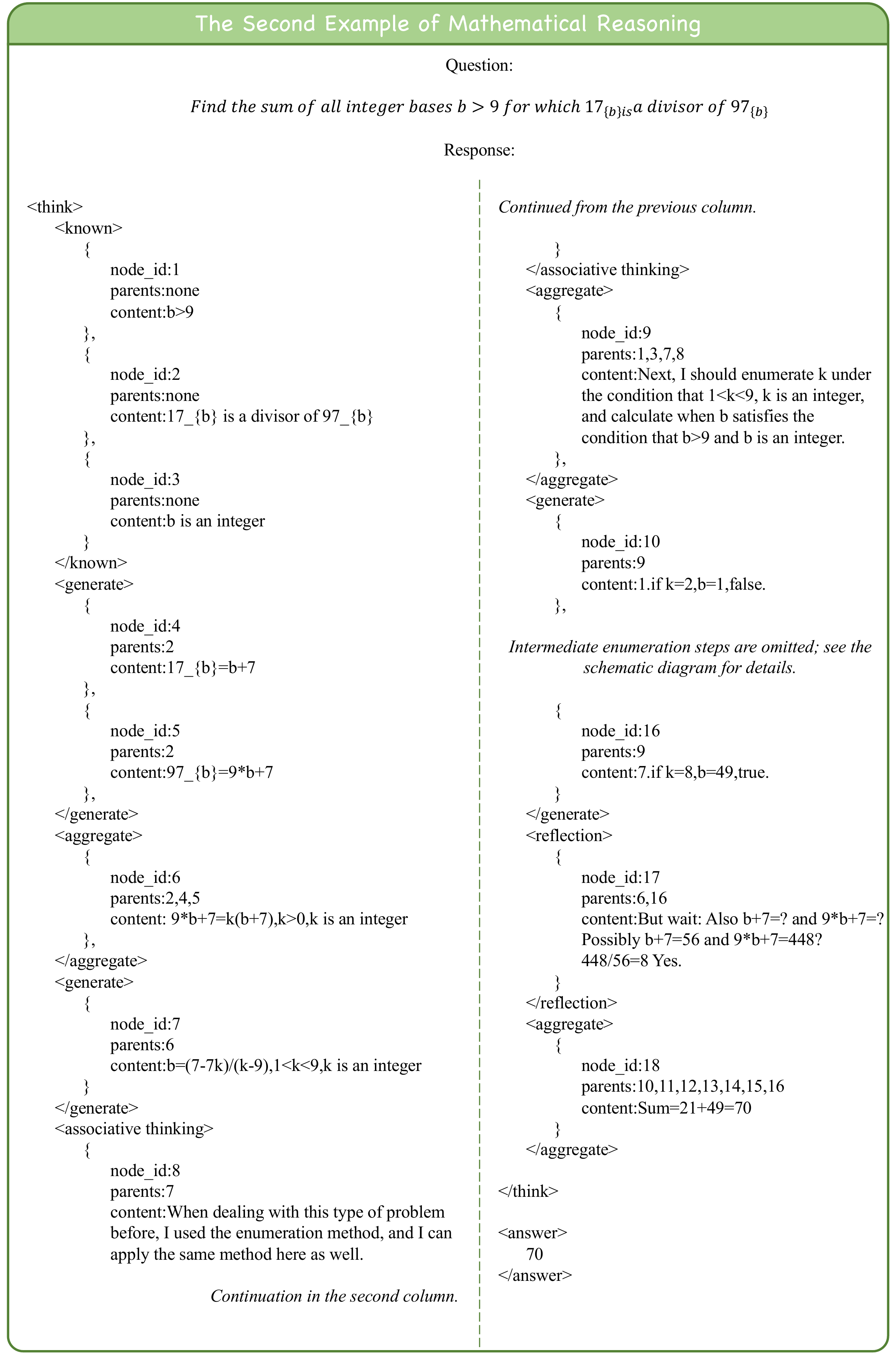}
    \caption{A mathematical example in graph-structured format.}
    \label{fig:math_example}
\end{figure*}

\begin{figure*}[htbp]
    \centering
    \includegraphics[width=\linewidth]{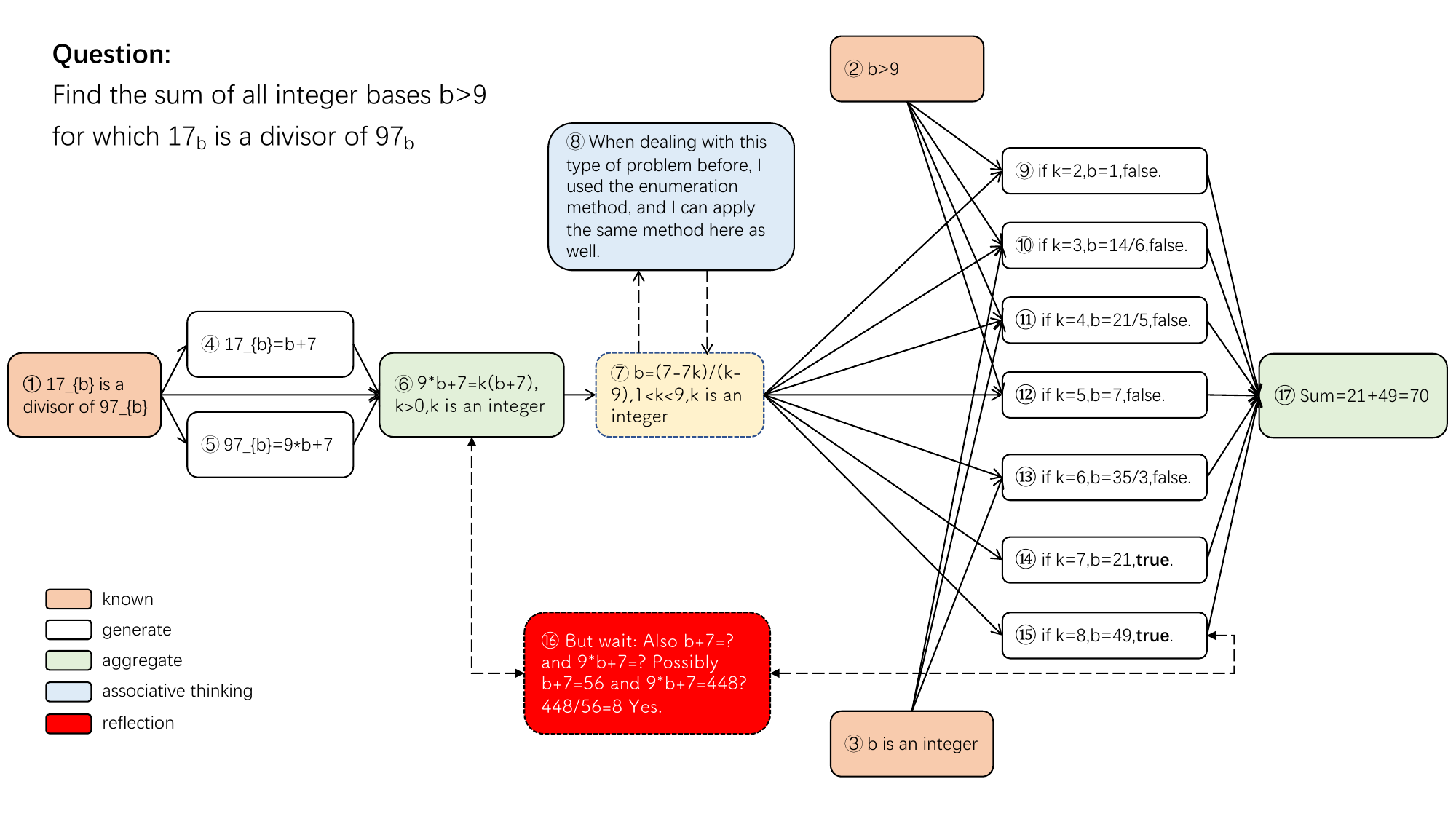}
    \caption{Graph visualization of the mathematical example.}
    \label{fig:math_example_graph}
\end{figure*}

We first present two mathematical examples that are originally included
in the graph-structured translation prompt.
The first example is designed to explicitly illustrate the use of
\emph{reverse thinking} and \emph{associative thinking}
(Figure~\ref{fig:math_example_reverse}).
The second example demonstrates another mathematical reasoning process
and is accompanied by a schematic visualization of its corresponding
reasoning graph, where nodes represent intermediate reasoning steps and
edges encode their dependency relations
(Figures~\ref{fig:math_example} and~\ref{fig:math_example_graph}).

\subsection{Code Generation Examples}
We include three code generation examples.
Each example corresponds to a representative question type of code generation task.

\begin{table*}[htbp]
    \centering
    \normalsize 
    \caption{Ablation Study Results of Length-Reducing Rewards.}
    \label{tab:ablation_length}
    \renewcommand{\arraystretch}{1} 
    \begin{tabular}{lcccccc}
        \toprule
        \multirow{2}{*}{\textbf{Method}} & \multicolumn{2}{c}{\textbf{MATH500}} & \multicolumn{2}{c}{\textbf{AMC23}} & \multicolumn{2}{c}{\textbf{AIME24}} \\
        \cmidrule(lr){2-3} \cmidrule(lr){4-5} \cmidrule(lr){6-7}
        & \textbf{Acc.} & \textbf{Length} & \textbf{Acc.} & \textbf{Length} & \textbf{Acc.} & \textbf{Length} \\
        \midrule
        PASC-GRPO                & 91.20 & 1260 & 82.50 & 1960 & 46.67 & 7040 \\
        w/o Connectivity         & 91.35 & 1850 & 81.55 & 2140 & 45.10 & 11200 \\
        w/o Conn. \& ERS Ratio & 90.52 & 1920 & 79.12 & 2483 & 43.25 & 22500 \\
        \bottomrule
    \end{tabular}
\end{table*}

\begin{figure*}[htbp]
    \centering
    \includegraphics[width=\textwidth]{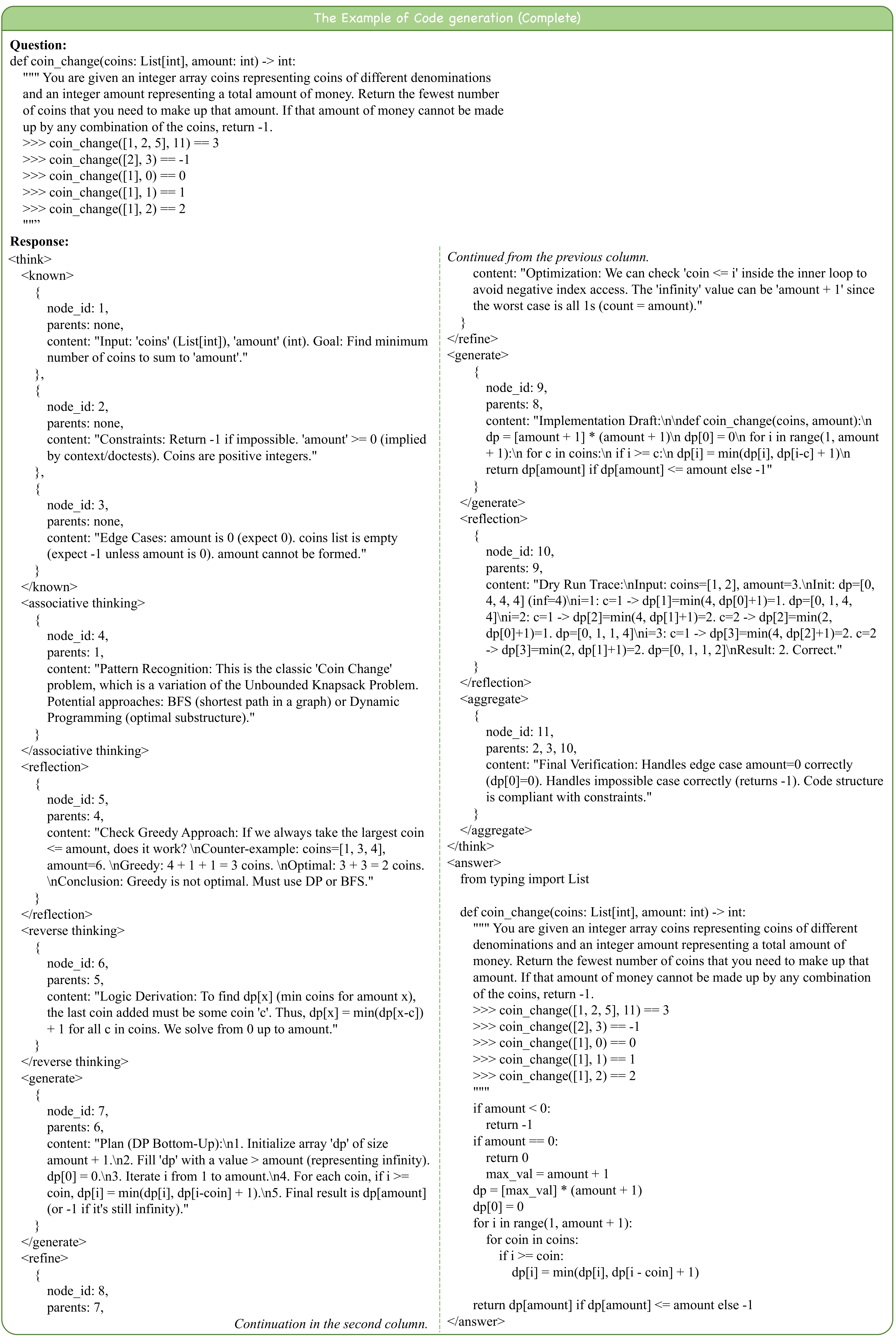}
    \caption{A code completion example.}
    \label{fig:code_complete}
\end{figure*}
\begin{figure*}[htbp]
    \centering
    \includegraphics[width=\textwidth]{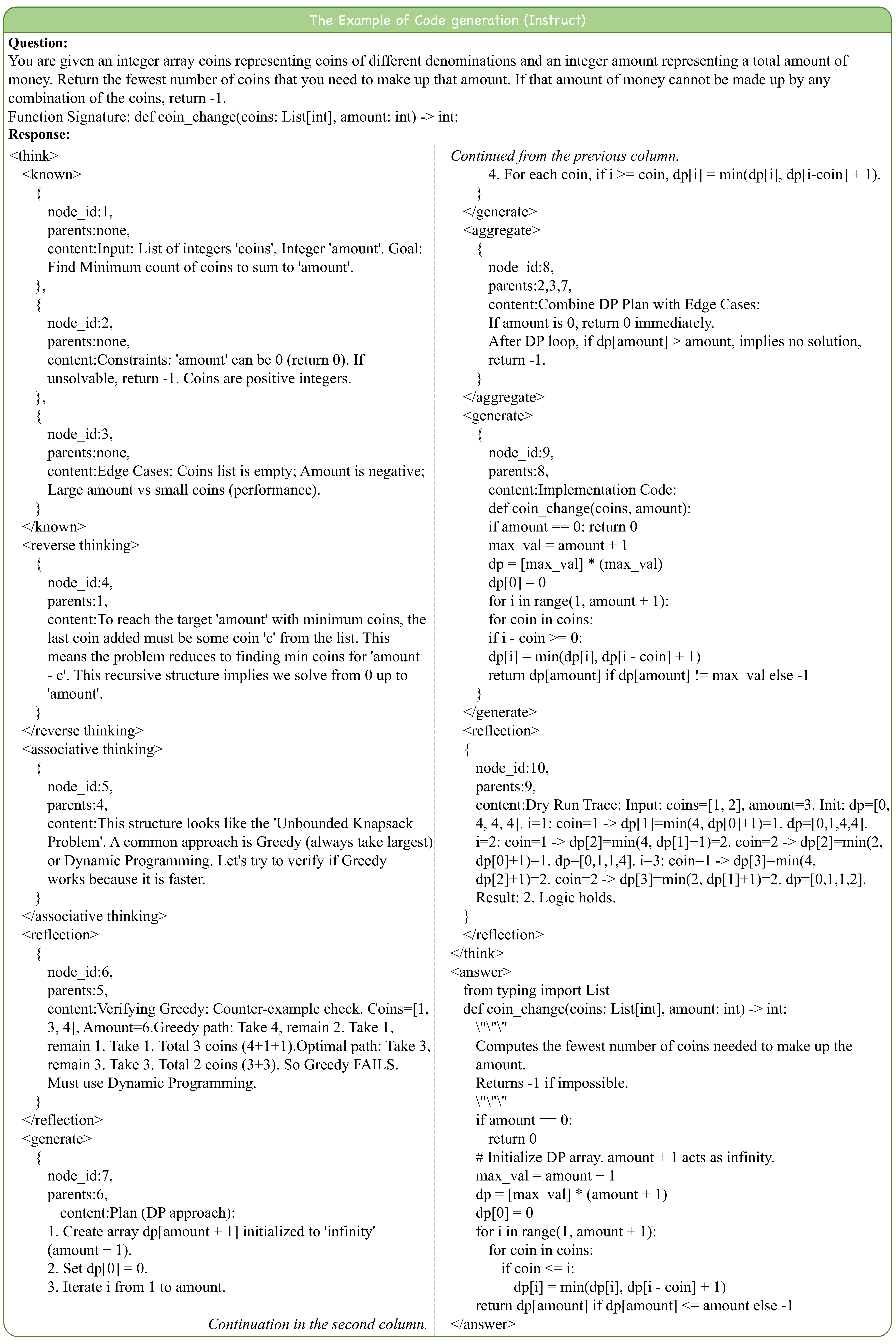}
    \caption{An instruction-following code generation example.}
    \label{fig:code_instruct}
\end{figure*}
\begin{figure*}[htbp]
    \centering
    \includegraphics[width=\textwidth]{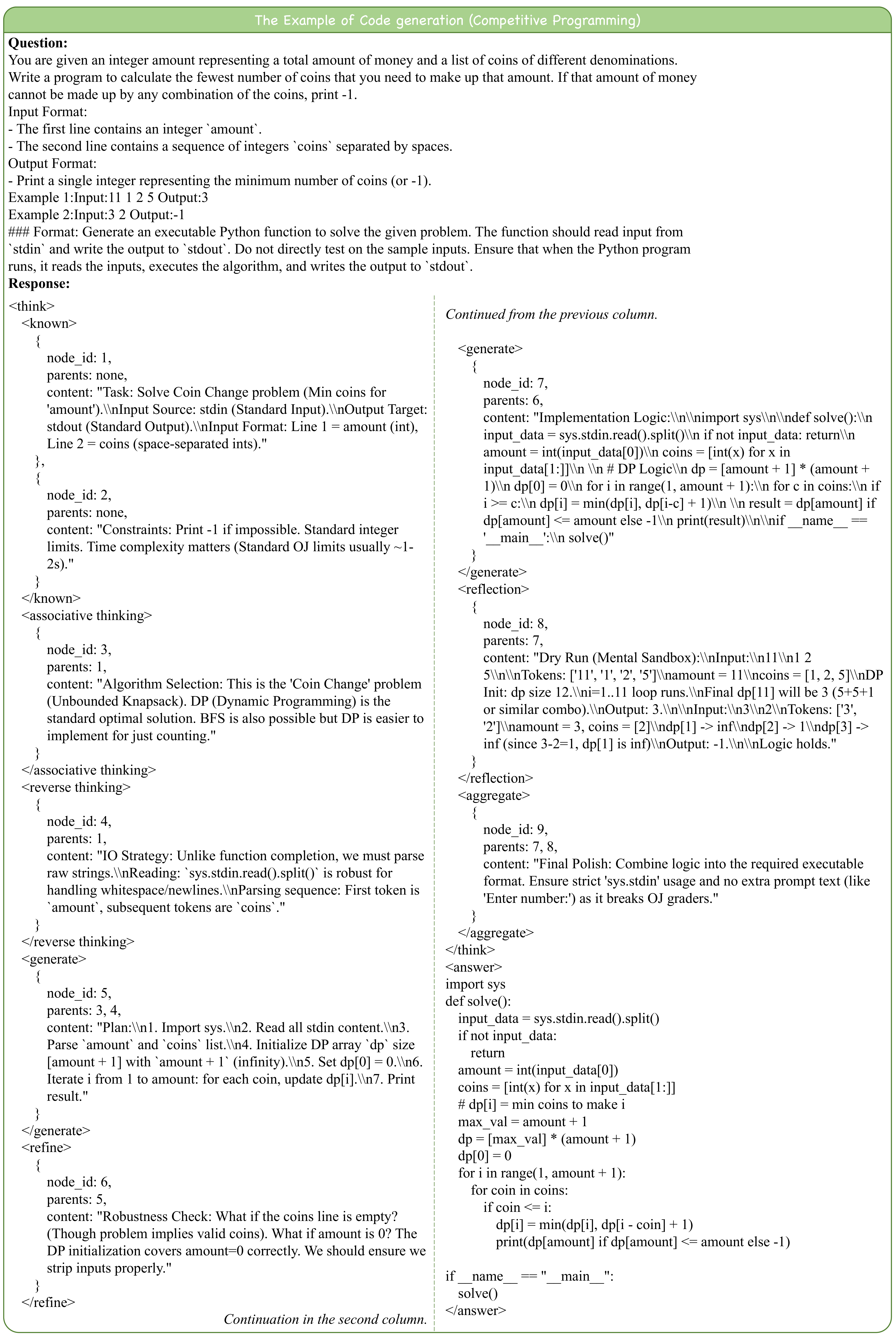}
    \caption{A competitive programming code generation example.}
    \label{fig:code_cp}
\end{figure*}

\begin{itemize}
    \item \textbf{Code Completion (Complete).}
    This type simulates the automatic code completion scenario in
    integrated development environments (IDEs).
    The model is provided with a function signature and a docstring as context
    and is required to generate the remaining function body.
    This task reflects the core capabilities of base code models such as
    Codex and StarCoder, primarily evaluating syntactic correctness and
    context-aware continuation, as exemplified by benchmarks like HumanEval
    (see Figure~\ref{fig:code_complete}).

    \item \textbf{Instruction Following (Instruct).}
    Also referred to as instruction-based or conversational code generation,
    this paradigm evaluates a model's alignment with human intent.
    The input consists of a natural language description of a programming task,
    and the model must generate a corresponding functional implementation.
    This setting is the dominant evaluation protocol for instruction-tuned
    models such as GPT-4 and DeepSeek-Coder-Instruct, emphasizing the
    translation of natural language semantics into executable logic
    (see Figure~\ref{fig:code_instruct}).

    \item \textbf{Competitive Programming (Online Judge).}
    This type emulates algorithmic contest environments such as ACM-style
    competitions or online judges (e.g., LeetCode).
    Unlike instruction-following tasks that generate a single function,
    competitive programming requires the model to produce a complete script
    that handles standard input and output streams.
    As a result, this paradigm is widely regarded as the most challenging
    form of code generation, demanding advanced algorithmic reasoning as well
    as robust handling of input parsing, output formatting, and edge cases
    (see Figure~\ref{fig:code_cp}).
\end{itemize}

\section{Graphical CoT Verification and Refinement Prompt}
\label{app:verify}

\begin{figure*}[htbp]
    \centering
    \includegraphics[width=\linewidth]{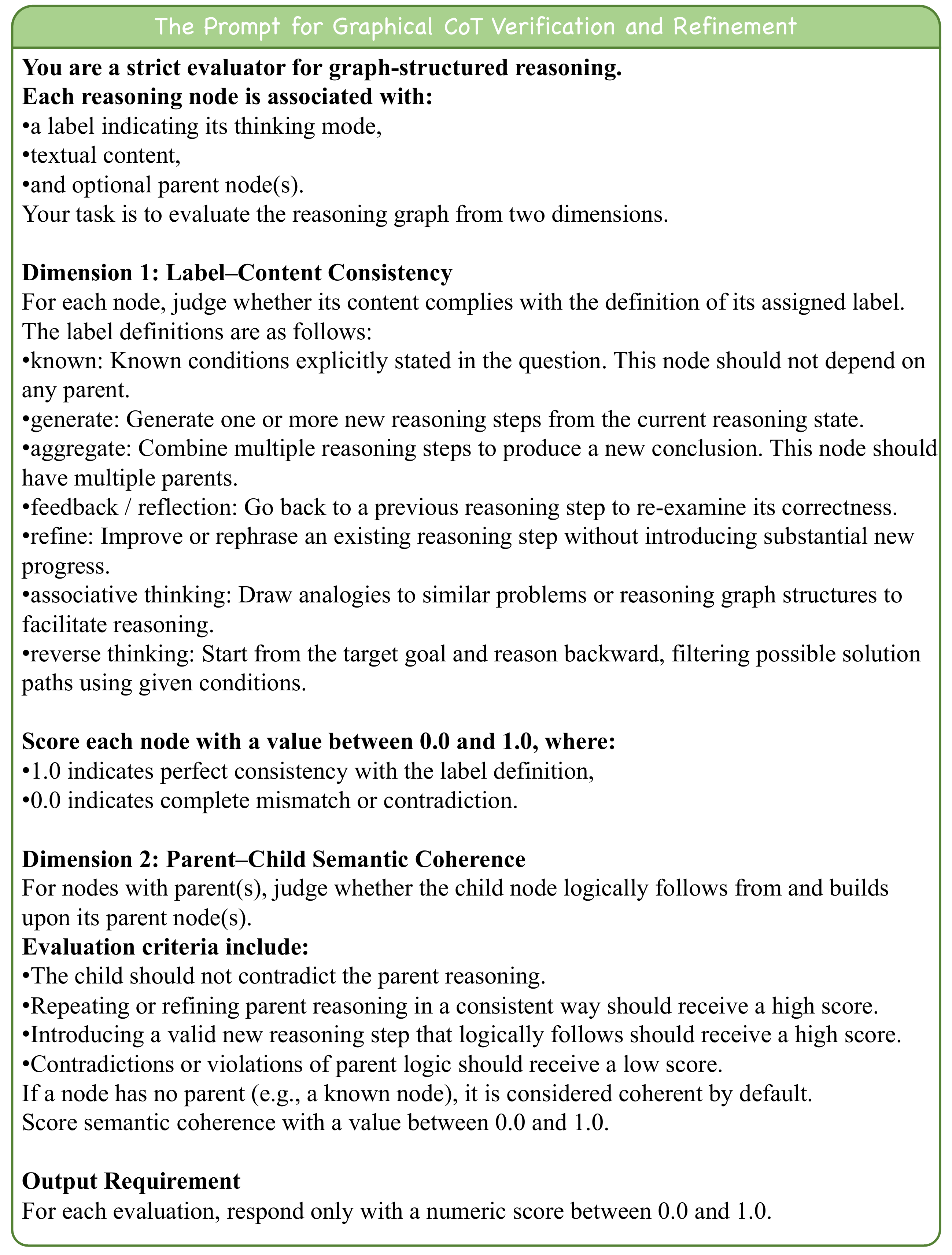}
    \caption{Prompt for ``Graphical CoT Verification and Refinement''.}
    \label{fig:prompt_verify}
\end{figure*}

This appendix presents the prompt for graphical Chain-of-Thought (CoT)
verification and refinement, which is shown in Figure~\ref{fig:prompt_verify}. All verification and refinement results are generated using the \texttt{Qwen3-Max} model.

\section{Training Data}
\label{app:training_data}

\subsection{Supervised Fine-Tuning (SFT) Data}
\begin{itemize}
    \item \textbf{Math Reasoning}.
    We use the full 40.3k samples from the DeepScaleR dataset.
    
    \item \textbf{Code Generation}.
    We curate 12k samples from the KodCode-V1-SFT-R1 subset,
    balancing task difficulty
    ($Easy:Middle:Hard = 4:3:3$)
    and task types
    ($Complete:Instruct:Online\_Judge = 4.5:4.5:1$).
\end{itemize}

\subsection{Reinforcement Learning (RL) Data}
\begin{itemize}
    \item \textbf{Math Reasoning}.
    We sample 5,000 problems from the DeepScaleR dataset.
    Problems are attempted four times by Qwen3-8B and stratified by difficulty
    based on success count.
    Problems failed in all attempts are discarded, followed by stratified
    sampling with a ratio of $1:1:1.5:1.5$.
    
    \item \textbf{Code Generation}.
    We directly employ the KodCode-Light-RL-10K subset.
\end{itemize}

\subsection{Data Leakage Check}
\label{app:data_leakage}

To prevent potential data leakage, we first collected all evaluation test sets relevant to our tasks.  
After selecting the training data for both supervised fine-tuning and reinforcement learning, we cross-checked each training sample against these test sets.  
Any overlapping instances would have been removed.  
Our inspection confirmed that the training sets contain no examples from the test sets, ensuring a clean separation between training and evaluation data.

\section{Ablation Study Details}
\label{app:ablation}

Table~\ref{tab:ablation_length} reports the detailed ablation results of
the proposed length-reducing rewards across multiple math reasoning benchmarks.
We progressively remove the connectivity reward and ERS information ratio,
to examine their individual contributions.
The results show that removing these components leads to a substantial
increase in reasoning length and a consistent degradation in accuracy,
especially on more challenging benchmarks such as AIME24,
highlighting their importance in controlling reasoning efficiency
without sacrificing correctness.

\end{document}